\RequirePackage{fix-cm}
\documentclass[smallextended]{svjour3}       
\smartqed  
\usepackage{graphicx}
\usepackage{amssymb}
\usepackage{algorithm,algpseudocode}
\usepackage{amsmath}
\usepackage[table]{xcolor}
\usepackage{graphics}
\usepackage{indentfirst}
\usepackage{times}
\usepackage{dsfont}
\usepackage{hyperref}
\usepackage{algorithm,algpseudocode}
\usepackage{multirow}
\usepackage{tabularx}
\usepackage[numbers,sort]{natbib}
\usepackage{setspace}
\usepackage{array,hhline}
\usepackage{pgf}

\begin{document}

\title{Feature selection algorithm based on Catastrophe model to improve the performance of regression analysis
}


\author{Mahdi Zarei 
}


\institute{Mahdi Zarei \at
University of California San Francisco,\\
              \email{mahdi.zarei@ucsf.edu}           
}

\date{}

\maketitle

\begin{abstract}
In this paper we introduce a new feature selection algorithm to remove the irrelevant or redundant features in the data sets. In this algorithm the importance of a feature is based on its fitting to the Catastrophe model. Akaike information criterion value is used for ranking the features in the data set. The proposed algorithm is compared with well-known RELIEF feature selection algorithm. Breast Cancer, Parkinson Telemonitoring data and Slice locality data sets are used to evaluate the model.
\end{abstract}

\section{Introduction}

Finding the informative features from a data is a complicated process. Many algorithms have been developed to remove the irrelevant features in the data set and improve the performance of analysis. For example multivariate feature selection statistics is used to reduce the complexity of the data analysis \cite{norman2006beyond}. Dimension reduction is another method to select informative features that many researchers applied to the features in the data \cite{ku2008comparison,o2007theoretical,mourao2006impact}. \par
In this paper, we introduce a new feature selection algorithm to improve performance of regression analysis.  Akaike information criterion value is used for ranking the features in the data set. The proposed algorithm is compared with well-known RELIEF feature selection algorithm. This algorithm is able to significantly reduce the number of features in this data set improving regression analysis accuracy.\par
Since our algorithm is based on the approaches from Catastrophe theory and Akaike information criterion, we start with a brief description of them.

\section{Cusp Catastrophe}

\begin{figure}[h]
\centering
\includegraphics[width=.32\textwidth]{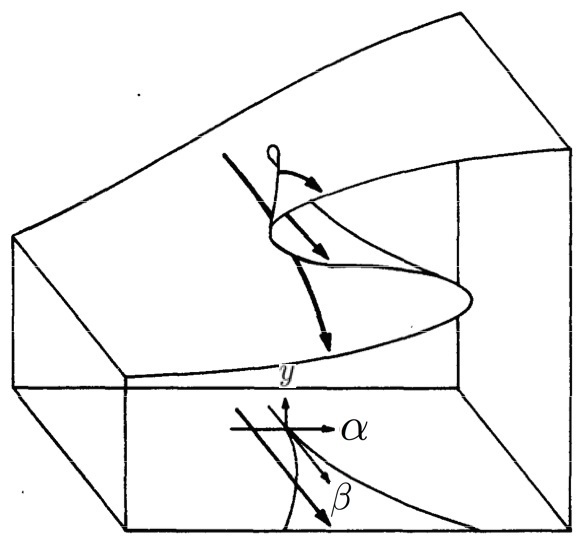}
\caption{Cusp surface \cite{cobb1980statistical}} \label{Cusp_Cobb}
\end{figure}

In this section we give a brief description of cusp model.
Consider the following dynamical system:
\begin{equation}\label{Cusp_01}
  \frac{\partial y}{\partial t}=-\frac{\partial V(y;c)}{\partial t}, y \in R^k, c \in R^p,
\end{equation}
where $V$ is the potential function, $y(t)$ represents the system's state variable(s), $c$ shows one or multiple (control) parameter(s) whose value(s) determine the specific structure of the system. If $y$ is at a point where
\begin{equation}\label{Cusp_02}
\frac{\partial V(y;c)}{\partial t}=0
\end{equation}
the system is in equilibrium. The function $V (y; c)$ acquires a minimum with respect to $y$ at a non-equilibrium point.
Equilibrium points that correspond to minima of $V (y; c)$ are stable equilibrium points because the system will return to such a point after a small perturbation to the system's state. The equilibrium points that correspond to maxima of $V (y; c)$ are unstable equilibrium points because a perturbation of the system's state will cause the system to move away from the equilibrium point towards a stable equilibrium point. Equilibrium points that correspond neither to maxima nor to minima of $V (y; c)$, at which the Hessian matrix ($\partial^2V (y)/ \partial y_i \partial y_j$) has eigenvalues equal to zero, are called degenerate equilibrium points. When the control variables of the system are changed. System can give rise to unexpected bifurcations in its equilibrium states at these points when the control variables of the system are changed \cite{saunders1980introduction,zeeman1977catastrophe,grasman2009fitting}. \\
Cusp model that is the simplest form of Catastrophe and can be formulated as follows:
\begin{equation}\label{Cusp_03}
-V(y;\alpha,\beta)=\alpha y +\frac{1}{2}\beta y^2-\frac{1}{4}y^4,
\end{equation}
where $V$ is the canonical form of the potential function for the Cusp model and its equilibrium points is a function of the control parameters $\alpha$ and $\beta$ (see Figure \ref{Cusp_Cobb}). The control parameters are the solution to the equation
\begin{equation}\label{Cusp_04}
\alpha+\beta y-y^3=0.
\end{equation}
This equation has one solution if $\delta=27 \alpha-4\beta^3$ that is greater than zero, and has three solutions if $\delta <0$ \cite{cobb1980estimation,grasman2009fitting}.

\section{Akaike information criterion}
Akaike information criterion (AIC) is a model quality measure for a given data \cite{akaike1974new,burnham2004multimodel}. For a model AIC measure can be defined as follow \cite{sakamoto1986akaike,bozdogan2000akaike}:
\begin{equation}
 AIC=-2log \mathrm{L} (\widehat{\theta})+2k,
\end{equation}
where $\mathrm{L} (\widehat{\theta})$ is the maximized  likelihood function and $k$ is the number of free parameters in the model. The smaller value of AIC shows that data is the better fit to model. In the proposed algorithm, we used the reverse value of AIC for ranking the features in our data.

\section{The feature selection algorithm}
\begin{figure}
\centering
\includegraphics[width=.8\textwidth]{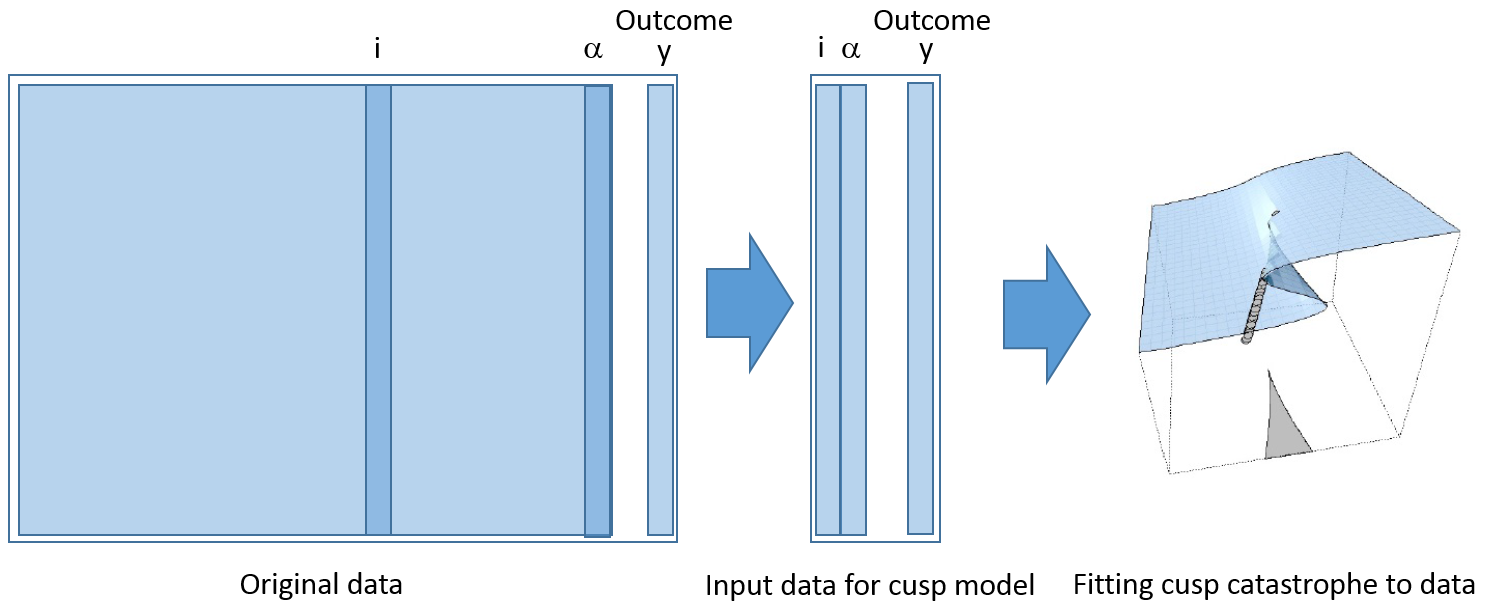}
\caption{Preparing input features for Cusp Catastrophe model}
\label{Cusp_InputData}
\end{figure}

In the Catastrophe theory, small change in certain parameters of a system can cause equilibria to appear or disappear \cite{thom1983mathematical,zeeman1977catastrophe}. We used this characteristic of the Catastrophe model to find the features that are more affective in regression analysis. In the proposed algorithm the features that better change the dynamic of outcome feature or features are considered as informative features. Assume that we are given a data set $A$ with $N$ features that $z$ is outcome feature. The algorithm takes each feature $i$
from the data set and considers it as bifurcation variable in the Cusp Catastrophe model. If this variable affects the dynamic of the system (outcome feature), it is the informative feature. The AIC value of the Cusp model is computed for each feature for ranking. The ranking of a feature $i$ can be formulated as follows:
\begin{equation}\label{cs_ranking_01}
  AIC_i=AIC(-V(y; \alpha ,i)),
\end{equation}
where $V$ is the potential function for the Cusp model (see Equation \ref{Cusp_03}), $AIC_i$ is the AIC value of the Cusp model for the feature $i$ as bifurcation value ($\beta$) and $\alpha$ is the asymmetric value in the Cusp model. Figure \ref{Cusp_InputData} shows the preparing the input parameters for Cusp model where the outcome feature is considered as the state variable and the features  $i$ and the last features are considered as bifurcation and asymmetric values, respectively. The state variable and control values can be computed as follows \cite{grasman2009fitting}:

\begin{align}\label{cusp_cobb_03}
  y[t] = w[0] + w[1] * Y[t,1] + ... + w[p] * Y[t,p], \\
  \alpha[t] = a[0] + a[1] * X[t,1] + ... + a[p] * X[t,p], \\
  \beta[t] = b[0] + b[1] * X[t,1] + ... + b[p] * X[t,p],
\end{align}
where $X[t,p]$'s are independent and $Y[t,p]$'s are dependent features in the data set. The vectors $a[j]$'s, $b[j]$'s and $w[j]$'s are estimated by means of maximum likelihood. The rank of each feature $i$ in the data set can be calculated as follows:

\begin{equation}\label{cs_ranking_02}
  rank_i \leftarrow \frac{1}{AIC_i}.
\end{equation}

More details about the model are shown in the Algorithm \ref{alg:Feature_selection_CS}.

\medskip
  \begin{algorithm}[H]
    \caption{Feature selection algorithm based on the Cusp Catastrophe model and AIC ranking}
    \label{alg:Feature_selection_CS}
    \begin{algorithmic}[1]
      \State(Initialization) $N \leftarrow$ Number of features ,$N_F \leftarrow$ Number of informative features , $\alpha \leftarrow feature_N$, $i \leftarrow 1$ and  $\alpha$ is asymmetric variable 
      \State \label{step:compute_beta} Let $\beta \leftarrow feature_i$ be bifurcation value in the Cusp model
      \State \label{step:cumpute_cusp} (Fitting the Cusp model using $\alpha$ and $\beta$) Let $AIC_i$ be the Akaike information criterion value of the fitting Cusp model using parameters $\alpha$ and $\beta$
      \State \label{step:ranking} (Ranking the feature) $rank_i \leftarrow \frac{1}{AIC_i}$ is the rank of feature $i$ in the dataset
      \State \label{step:delet_irrelevant} if $\frac{1}{AIC_i} \le t$ then $feature_i$ is not informative and eliminate it, $i \leftarrow i+1$ and go to \ref{step:stopping_criterion}
      \State \label{step:stopping_criterion} (Stopping criterion) if $i > N_F$ stop. Otherwise go to Step \ref{step:compute_beta}
      \State \label{step:Returning_result} (Retraining informative features) Return $N_F$ informative features.
    \end{algorithmic}
  \end{algorithm}
Here $N$ is the number of all feature in the data set and $N_F$ ($N_F < N$) is the number of informative features. For all features $i$ of the data set their rank in the data set is computed ($rank_i$). The set of informative features with $N_F$ features is the outcome of the algorithm.


\section{RELIEF feature selection algorithm}
Next, we give a brief description of the RELIEF algorithm. More detailed description can be found in \cite{Kira1992,Kononenko1994,Robnik-Sikonja1997}. For a given data set with $m$ samples, and threshold of relevancy $\tau$ ($ 0 \leq \tau \leq 1 $), it detects those features which are statistically relevant to the target concept ($Y=f(X)$). Differences of feature value between two instances $X$ and $Y$ are defined by the following function $diff$ \cite{kira1992feature}. \\

\begin{equation}
  diff(x_k,y_k)=(x_k-y_k)/nu_k,
\end{equation}
where $nu_k$ is a normalization unit to normalize the values of $diff$ into the interval $[0, l]$. RELIEF picks a sample composed of $m$ triplets of an instance $X$, it's same-class instance ($nearHit$) and closest different-class instance ($nearMiss$). RELIEF uses the $p$-dimensional Euclidean distance for selecting $nearHit$ and $nearMiss$. In every routine the feature weight $W$ vector is updated as follows:
\begin{equation} \label{RELIEF_W}
  W_i=W_{i-1}-(x_i-nearHit_i)^2+(x_i-nearMiss_i)^2.
\end{equation}
Then the average feature weight vector relevance is determined for every sample triple. Finally, it chooses the features whose average weight is above the given threshold $\tau$.


\section{Experimental results}
The effectiveness of the proposed algorithm is verified using three different data sets: Parkinson's Telemonitoring, Breast Cancer and Slice locality from UCI machine learning repository \cite{blake1998uci}. Numerical experiments have been carried out on a PC with Processor Intel(R) Core(TM) i5-3470S CPU 2.90 GHz and 8 GB RAM running under Windows 7. \\
In numerical experiments we apply the proposed algorithm to find a ranking sequence of features in data sets. Then we apply different regression analysis algorithms from WEKA to compute regression error with subsets of features. The following regression analysis algorithms from WEKA are used in numerical experiments:


\begin{itemize}
  \item Linear regression: Linear regression finds the best curve to fit the data by computing the relationship between a scalar dependent variable $y$ and one or more explanatory variables denoted $X$. It applies least squares, which minimizes the sum of the distance from the line for each of points. The actual observations, $y_i$, may be slightly off the population line because of variability in the population. The equation is $y_i = \beta_0 + \beta_1 x_i +\epsilon_i$, where $\epsilon_i$ is the deviation from the population line which is called the residual \cite{barlow1993numerical,neter1983applied}.
  \item K nearest neighbors regressor: The algorithm computes the mean of the function values of its $K$-nearest neighbours \cite{kramer2011unsupervised}.

  \item M5Rulles: It generates rules for numeric prediction by separate-and-conquer and at each iteration builds a model tree using M5 and makes the "best" leaf into a rule \cite{Holmes1999,Quinlan1992,Wang1997}
  \item REPTree: Reptree is a fast tree learner that uses reduced error pruning \cite{witten2005data}.
\end{itemize}

\subsection{Results for Breast cancer data set}
Breast Cancer Wisconsin (Prognostic) Data Set contains 30 features with 569 samples. Each record represents follow-up data for one breast cancer case \cite{mangasarian1995breast,street1995inductive} . Table \ref{breast_cancer_fs} presents the error of analysing the data using for regression analysis algorithms. The second row shows the number of features before and after feature selection. Results from this table demonstrate that features selected by the proposed algorithm allow us to reduce the mean absolute error (MAE) regression. MAE is calculated as follows:

\begin{equation}\label{CS_MAE}
  MAE= \frac{1}{n}\sum_{i=1}^{n}\left | f_i-y_i \right |,
\end{equation}
where $n$ is the number of observation,  $f_i$ is the predicted and $y_i$ is the true values. Although this data set is not noisy the proposed algorithm is able to significantly reduce the number of features without deteriorating the regression error. Regression errors with the subsets of features which are better than that of for all features are presented in bold font.


\begin{table}
\centering
\small
\caption{Performance of regression analysis algorithms for breast cancer data set}
\begin{tabular}{c|c|cccccc}

\hline
                            & \textbf{Original data} & \multicolumn{6}{c}{\textbf{After feature selection}}                                                                                                                                          \\ \hline
Number of features & 30            & \multicolumn{1}{c|}{25} & \multicolumn{1}{c|}{20} & \multicolumn{1}{c|}{15} & \multicolumn{1}{c|}{10} & \multicolumn{1}{c|}{6} & 5      \\ \hline
Linear Regression	&	0.003	&	\textbf{0.003}	&	 \textbf{0.003}	&	 \textbf{0.003}	&	\textbf{0.003}	&	 \textbf{0.003}	&	0.004	\\ \hline
IBK	&	0.008	&	\textbf{0.008}	&	\textbf{0.008}	&	 \textbf{0.007}	&	\textbf{0.007}	 &	\textbf{0.006}	&	 \textbf{0.007}	\\ \hline
M5P	&	0.003	&	\textbf{0.003}	&	\textbf{0.003}	&	 \textbf{0.003}	&	\textbf{0.003}	 &	\textbf{0.003}	&	 0.004	\\ \hline
M5Rules	&	0.003	&	\textbf{0.003}	&	\textbf{0.003}	&	 \textbf{0.003}	&	 \textbf{0.003}	&	\textbf{0.003}	&	 0.004	\\ \hline
\end{tabular}
\label{breast_cancer_fs}
\end{table}

\subsection{Results for Slice locality data set}
Slice locality data set consists of 384 features extracted from 53500 CT images. The CT images are from 74 different patients (43 male, 31 female). The class variable of this data set is the location of the CT slice on the axial axis of the human body \cite{graf20112d}. This data set is available on UCI Machine Learning Repository.\par
Results for 10 subjects of Slice locality data set are presented in Tables \ref{Slice_localization_IBK}-\ref{Slice_localization_M5Rules}. In these tables regression error obtained by regression algorithms are given. The second line in all tables contains a number of features of original data and after feature selection. Table \ref{Slice_localization_IBK} presents results for all subjects using IBK algorithm. One can see that the IBK algorithm achieved the better accuracy for all subjects data set except subject number 10 using 380 features.
Table \ref{Slice_localization_LR} presents results for all subjects using Logistic regression algorithm. The use of the proposed algorithm allows improving the performance of Logistic regression using 250 features for Subject 1 and 150 features for Subjects 2 and 3. The best performance for Subject 5 achieved using 100 features. Results are almost the same for other Subjects.\par
Tables \ref{Slice_localization_M5P} and \ref{Slice_localization_M5Rules} show results for all patients using M5P and M5Rules algorithms, respectively. Results for these two algorithms are very similar and one can see that the proposed algorithm can improve the accuracy of regression algorithms.

\begin{table}[h]
\centering
\footnotesize
\caption{IBK algorithm performance for 10 subjects from Slice locality data}
\begin{tabular}{c|c|ccccccc}
\hline
\textbf{}          & \textbf{Original data} & \multicolumn{7}{c}{\textbf{After feature selection}}                                                                                                                     \\ \hline
Number of features & 385                    & \multicolumn{1}{c|}{380} & \multicolumn{1}{c|}{350} & \multicolumn{1}{c|}{300} & \multicolumn{1}{c|}{250} & \multicolumn{1}{c|}{200} & \multicolumn{1}{c|}{150} & 100    \\ \hline
Patient1	&	0.059	&	\textbf{0.059}	&	\textbf{0.059}	 &	0.060	&	 0.061	&	0.065	&	0.063	&	0.083	\\ \hline
Patient2	&	0.080	&	\textbf{0.080}	&	0.081	&	 0.081	&	 0.082	&	0.083	&	0.085	&	0.103	\\ \hline
Patient3	&	0.076	&	\textbf{0.076}	&	\textbf{0.076}	 &	\textbf{0.075}	&	 \textbf{0.076}	&	0.077	&	0.086	 &	0.115	\\ \hline
Patient4	&	0.060	&	\textbf{0.060}	&	\textbf{0.060}	 &	0.061	&	 0.062	&	0.063	&	0.066	&	0.081	\\ \hline
Patient5	&	0.078	&	\textbf{0.078}	&	\textbf{0.078}	 &	0.079	&	 0.080	&	0.088	&	0.086	&	0.090	\\ \hline
Patient6	&	0.349	&	\textbf{0.349}	&	\textbf{0.349}	 &	\textbf{0.349}	&	 \textbf{0.336}	&	\textbf{0.346}	&	 0.456	&	0.466	\\ \hline
Patient7	&	0.081	&	\textbf{0.081}	&	\textbf{0.081}	 &	\textbf{0.081}	&	 \textbf{0.081}	&	0.087	&	0.091	 &	0.099	\\ \hline
Patient8	&	0.087	&	\textbf{0.087}	&	\textbf{0.087}	 &	\textbf{0.087}	&	 \textbf{0.086}	&	\textbf{0.086}	&	 0.093	&	0.099	\\ \hline
Patient9	&	0.364	&	\textbf{0.364}	&	0.370	&	 0.370	&	 0.364	&	0.380	&	0.494	&	0.516	\\ \hline
Patient10	&	0.098	&	\textbf{0.098}	&	0.100	&	 0.104	&	 0.103	&	0.105	&	0.110	&	0.139	\\ \hline
\end{tabular}
\label{Slice_localization_IBK}
\end{table}

\begin{table}[h]
\centering
\footnotesize
\caption{Logistic regression algorithm performance for 10 subjects from Slice locality data}
\begin{tabular}{c|c|ccccccc}
\hline
\textbf{}          & \textbf{Original data} & \multicolumn{7}{c}{\textbf{After feature selection}}                                                                                                                              \\ \hline
Number of features & 385                    & \multicolumn{1}{c|}{380} & \multicolumn{1}{c|}{350} & \multicolumn{1}{c|}{300} & \multicolumn{1}{c|}{250} & \multicolumn{1}{c|}{200} & \multicolumn{1}{c|}{150} & 100             \\ \hline
Patient1	&	0.354	&	0.392	&	\textbf{0.250}	&	 \textbf{0.267}	&	 \textbf{0.284}	&	\textbf{0.326}	&	 0.411	&	0.570	\\ \hline
Patient2	&	0.496	&	\textbf{0.435}	&	\textbf{0.398}	 &	\textbf{0.367}	&	 \textbf{0.332}	&	\textbf{0.309}	&	 \textbf{0.376}	&	0.621	\\ \hline
Patient3	&	0.258	&	\textbf{0.256}	&	0.266	&	 \textbf{0.228}	&	 \textbf{0.226}	&	\textbf{0.226}	&	 \textbf{0.247}	&	0.361	\\ \hline
Patient4	&	0.282	&	0.294	&	0.305	&	 \textbf{0.281}	&	 0.294	&	\textbf{0.269}	&	0.373	&	 0.476	\\ \hline
Patient5	&	0.928	&	1.742	&	2.413	&	 \textbf{0.512}	&	 \textbf{0.440}	&	\textbf{0.469}	&	 \textbf{0.572}	&	\textbf{0.529}	\\ \hline
Patient6	&	0.435	&	0.439	&	0.456	&	0.440	&	 0.456	&	0.572	&	2.232	&	1.514	\\ \hline
Patient7	&	0.515	&	\textbf{0.500}	&	\textbf{0.460}	 &	\textbf{0.426}	&	 \textbf{0.420}	&	\textbf{0.414}	&	 \textbf{0.443}	&	0.756	\\ \hline
Patient8	&	1.306	&	\textbf{1.272}	&	\textbf{1.275}	 &	\textbf{1.275}	&	 1.449	&	\textbf{1.234}	&	1.457	 &	2.025	\\ \hline
Patient9	&	0.549	&	\textbf{0.539}	&	0.567	&	 0.532	&	 \textbf{0.497}	&	0.860	&	1.857	&	7.839	 \\ \hline
Patient10	&	0.570	&	\textbf{0.565}	&	\textbf{0.513}	 &	\textbf{0.522}	&	 \textbf{0.508}	&	\textbf{0.492}	&	 \textbf{0.506}	&	0.681	\\ \hline
\end{tabular}
\label{Slice_localization_LR}
\end{table}

\begin{table}[h]
\centering
\footnotesize
\caption{M5P algorithm performance for 10 subjects from Slice localization data}
\begin{tabular}{c|c|ccccccc}
\hline
\textbf{}          & \textbf{Original data} & \multicolumn{7}{c}{\textbf{After feature selection}}                                                                                                                              \\ \hline
Number of features & 385                    & \multicolumn{1}{c|}{380} & \multicolumn{1}{c|}{350} & \multicolumn{1}{c|}{300} & \multicolumn{1}{c|}{250} & \multicolumn{1}{c|}{200} & \multicolumn{1}{c|}{150} & 100             \\ \hline
Patient1	&	0.299	&	\textbf{0.299}	&	0.301	&	 \textbf{0.297}	&	 \textbf{0.294}	&	\textbf{0.293}	&	 \textbf{0.298}	&	0.338	\\ \hline
Patient2	&	0.455	&	\textbf{0.455}	&	\textbf{0.440}	 &	\textbf{0.443}	&	 \textbf{0.441}	&	0.471	&	 \textbf{0.451}	&	\textbf{0.452}	\\ \hline
Patient3	&	0.352	&	\textbf{0.352}	&	\textbf{0.352}	 &	\textbf{0.349}	&	 0.358	&  \textbf{0.343}	&	 \textbf{0.342}	&	\textbf{0.337}	\\ \hline
Patient4	&	0.341	&	0.347	&	0.348	&	0.350	&	 \textbf{0.339}	&	\textbf{0.310}	&	\textbf{0.319}	&	 \textbf{0.325}	\\ \hline
Patient5	&	0.458	&	\textbf{0.458}	&	\textbf{0.427}	 &	\textbf{0.404}	&	 \textbf{0.395}	&	\textbf{0.375}	&	 \textbf{0.385}	&	\textbf{0.396}	\\ \hline
Patient6	&	1.334	&	\textbf{1.297}	&	\textbf{1.289}	 &	\textbf{1.326}	&	 1.357	&	1.136	&	\textbf{1.229}	 &	\textbf{1.291}	\\ \hline
Patient7	&	0.472	&	\textbf{0.467}	&	\textbf{0.472}	 &	\textbf{0.472}	&	 \textbf{0.469}	&	0.476	&	0.475	 &	0.490	\\ \hline
Patient8	&	0.782	&	0.797	&	0.801	&	0.801	&	 \textbf{0.720}	&	\textbf{0.744}	&	\textbf{0.728}	&	 \textbf{0.728}	\\ \hline
Patient9	&	1.214	&	\textbf{1.214}	&	\textbf{1.175}	 &	\textbf{1.189}	&	 \textbf{1.152}	&	\textbf{1.020}	&	 1.683	&	1.754	\\ \hline
Patient10	&	0.561	&	\textbf{0.546}	&	\textbf{0.542}	 &	\textbf{0.513}	&	 \textbf{0.513}	&	\textbf{0.519}	&	 \textbf{0.509}	&	\textbf{0.519}	\\ \hline

\end{tabular}
\label{Slice_localization_M5P}
\end{table}

\begin{table}[h]
\centering
\footnotesize
\caption{M5Rules algorithm performance for 10 subjects for Slice localization data}
\begin{tabular}{c|c|ccccccc}
\hline
\textbf{}          & \textbf{Original data} & \multicolumn{7}{c}{\textbf{After feature selection}}                                                                                                                              \\ \hline
Number of features & 385                    & \multicolumn{1}{c|}{380} & \multicolumn{1}{c|}{350} & \multicolumn{1}{c|}{300} & \multicolumn{1}{c|}{250} & \multicolumn{1}{c|}{200} & \multicolumn{1}{c|}{150} & 100             \\ \hline
Patient1	&	0.331	&	0.319	&	\textbf{0.313}	&	 0.368	&	 0.370	&	\textbf{0.322}	&	\textbf{0.272}	&	 2.217	\\ \hline
Patient2	&	0.455	&	\textbf{0.455}	&	\textbf{0.360}	 &	\textbf{0.339}	&	 \textbf{0.347}	&	0.557	&	 \textbf{0.445}	&	0.490	\\ \hline
Patient3	&	0.508	&	\textbf{0.508}	&	\textbf{0.508}	 &	\textbf{0.477}	&	 \textbf{0.432}	&	\textbf{0.413}	&	 \textbf{0.388}	&	\textbf{0.420}	\\ \hline
Patient4	&	0.328	&	\textbf{0.307}	&	\textbf{0.311}	 &	\textbf{0.328}	&	 0.333	&	0.294	&	\textbf{0.309}	 &	\textbf{0.317}	\\ \hline
Patient5	&	0.481	&	\textbf{0.479}	&	\textbf{0.410}	 &	0.507	&	 0.508	&	\textbf{0.458}	&	0.492	&	 \textbf{0.412}	\\ \hline
Patient6	&	1.562	&	\textbf{1.320}	&	\textbf{1.231}	 &	\textbf{1.313}	&	 \textbf{1.338}	&	\textbf{1.030}	&	 \textbf{1.480}	&	\textbf{1.242}	\\ \hline
Patient7	&	0.783	&	\textbf{0.783}	&	0.784	&	 \textbf{0.783}	&	 \textbf{0.559}	&	\textbf{0.500}	&	 \textbf{0.412}	&	\textbf{0.611}	\\ \hline
Patient8	&	0.686	&	\textbf{0.687}	&	0.696	&	 0.696	&	 0.853	&	0.822	&	0.755	&	2.506	\\ \hline
Patient9	&	1.476	&	\textbf{1.476}	&	\textbf{1.220}	 &	\textbf{1.249}	&	 \textbf{1.162}	&	\textbf{1.260}	&	 \textbf{0.968}	&	1.952	\\ \hline
Patient10	&	0.815	&	\textbf{0.693}	&	\textbf{0.727}	 &	\textbf{0.714}	&	 \textbf{0.688}	&	-	&	1.926	&	 0.586	\\ \hline
\end{tabular}
\label{Slice_localization_M5Rules}
\end{table}

\subsection{Results for Parkinsons Telemonitoring data set}
In this paper, we present the results for Parkinsons Telemonitoring data set. This data set composed of a range of biomedical voice measurements from 42 people with early-stage Parkinson's disease. Here we analyzed 15 subjects from this data set.
Results for subjects of Parkinsons Telemonitoring data set are presented in Tables \ref{Parkinson_IBK}-\ref{Parkinson_M5Rulles}. This is illustration of a number of features in original data and after feature selection. The number of features in original data is 18. \\
Table \ref{Parkinson_IBK} shows the results for the error of the data using IBK regressor algorithm. The use of a very small subset of features can  provide better performance for almost all subjects.
Table \ref{Parkinson_LR} presents the results for Logistic regression algorithm. The proposed algorithm can reduce the error of more than 70\% of cases.
The situation is almost the same for the M5P algorithm \ref{Parkinson_M5P}, but M5Rulles algorithm provides better performance and the accuracy is increased for all subjects except Subjects 14 and 15.

\begin{table}[h]
\caption{IBK algorithm performance for Parkinson's disease data}
\scriptsize
\begin{tabular}{c|c|cccccccc}
\hline
\textbf{}                   & \textbf{Original  data} & \multicolumn{8}{c}{\textbf{After feature selection}}                                                                                                                                                                                                                            \\ \hline
\textbf{Number of features} & \textbf{18}            & \multicolumn{1}{c|}{\textbf{11}} & \multicolumn{1}{c|}{\textbf{10}} & \multicolumn{1}{c|}{\textbf{9}} & \multicolumn{1}{c|}{\textbf{8}} & \multicolumn{1}{c|}{\textbf{7}} & \multicolumn{1}{c|}{\textbf{6}} & \multicolumn{1}{c|}{\textbf{5}} & \multicolumn{1}{c|}{\textbf{4}} \\ \hline
Subject1	&	0.037	&	0.038	&	\textbf{0.037}	&	 0.038	&	 0.038	&	0.040	&	0.044	&	0.041	&	 0.042	\\ \hline
Subject2	&	0.039	&	\textbf{0.037}	&	\textbf{0.039}	 &	\textbf{0.038}	&	 \textbf{0.039}	&	0.040	&	 \textbf{0.036}	&	0.040	&	0.042	\\ \hline
Subject3	&	0.030	&	\textbf{0.027}	&	\textbf{0.027}	 &	\textbf{0.027}	&	 \textbf{0.027}	&	\textbf{0.026}	&	 \textbf{0.029}	&	\textbf{0.027}	&	\textbf{0.027}	\\ \hline
Subject4	&	0.039	&	\textbf{0.034}	&	\textbf{0.034}	 &	\textbf{0.035}	&	 \textbf{0.035}	&	\textbf{0.036}	&	 \textbf{0.034}	&	\textbf{0.035}	&	\textbf{0.037}	\\ \hline
Subject5	&	0.037	&	\textbf{0.033}	&	\textbf{0.032}	 &	\textbf{0.032}	&	 \textbf{0.030}	&	\textbf{0.029}	&	 \textbf{0.029}	&	\textbf{0.030}	&	\textbf{0.031}	\\ \hline
Subject6	&	0.034	&	0.037	&	\textbf{0.035}	&	 \textbf{0.033}	&	 \textbf{0.033}	&	\textbf{0.033}	&	 \textbf{0.031}	&	\textbf{0.031}	&	\textbf{0.031}	\\ \hline
Subject7	&	0.040	&	\textbf{0.033}	&	\textbf{0.033}	 &	\textbf{0.034}	&	 \textbf{0.030}	&	\textbf{0.034}	&	 \textbf{0.035}	&	\textbf{0.036}	&	\textbf{0.035}	\\ \hline
Subject8	&	0.032	&	\textbf{0.031}	&	0.033	&	 0.034	&	 \textbf{0.032}	&	0.033	&	0.036	&	0.036	 &	0.036	\\ \hline
Subject9	&	0.041	&	\textbf{0.038}	&	\textbf{0.038}	 &	\textbf{0.039}	&	 \textbf{0.039}	&	\textbf{0.037}	&	 \textbf{0.036}	&	\textbf{0.036}	&	\textbf{0.039}	\\ \hline
Subject10 & 0.044 & 0.037          & 0.039          & 0.039          & 0.044          & 0.046          & 0.044          & 0.042          & 0.040          \\ \hline
Subject11 & 0.022 & \textbf{0.022} & \textbf{0.022} & \textbf{0.021} & \textbf{0.021} & \textbf{0.021} & \textbf{0.020} & \textbf{0.021} & 0.023          \\ \hline
Subject12 & 0.030 & \textbf{0.024} & \textbf{0.024} & \textbf{0.028} & \textbf{0.029} & \textbf{0.030} & 0.032          & 0.032          & \textbf{0.030} \\ \hline
Subject13 & 0.040 & 0.042          & 0.044          & 0.042          & 0.047          & \textbf{0.039} & 0.051          & 0.049          & 0.049          \\ \hline
Subject14 & 0.032 & \textbf{0.030} & \textbf{0.030} & \textbf{0.031} & \textbf{0.031} & \textbf{0.031} & \textbf{0.031} & 0.033          & 0.033          \\ \hline
Subject15 & 0.032 & \textbf{0.031} & \textbf{0.030} & \textbf{0.030} & \textbf{0.032} & \textbf{0.031} & \textbf{0.030} & \textbf{0.032} & \textbf{0.032}\\
 \hline
\end{tabular}
\label{Parkinson_IBK}
\end{table}

\begin{table}[h]
\caption{Linear regression algorithm performance for Parkinson's disease data}
\scriptsize
\begin{tabular}{c|c|cccccccc}
\hline
\textbf{}                   & \textbf{Original data} & \multicolumn{8}{c}{\textbf{After feature selection}}                                                                                                                                                                                                            \\ \hline
\textbf{Number of features} & \textbf{18}            & \multicolumn{1}{c|}{\textbf{11}} & \multicolumn{1}{c|}{\textbf{10}} & \multicolumn{1}{c|}{\textbf{9}} & \multicolumn{1}{c|}{\textbf{8}} & \multicolumn{1}{c|}{\textbf{7}} & \multicolumn{1}{c|}{\textbf{6}} & \multicolumn{1}{c|}{\textbf{5}} & \textbf{4}      \\ \hline
Subject1  & 0.030 & \textbf{0.028} & \textbf{0.028} & \textbf{0.028} & \textbf{0.028} & \textbf{0.028} & \textbf{0.028} & \textbf{0.029} & \textbf{0.029} \\ \hline
Subject2  & 0.028 & \textbf{0.028} & 0.030          & 0.031          & 0.030          & 0.030          & 0.030          & 0.030          & 0.030          \\ \hline
Subject3  & 0.018 & 0.019          & \textbf{0.018} & 0.020          & 0.020          & 0.020          & 0.022          & 0.021          & 0.021          \\ \hline
Subject4  & 0.029 & \textbf{0.028} & \textbf{0.027} & \textbf{0.027} & \textbf{0.027} & \textbf{0.027} & \textbf{0.027} & \textbf{0.027} & \textbf{0.028} \\ \hline
Subject5  & 0.024 & 0.025          & 0.025          & 0.025          & 0.025          & 0.026          & 0.026          & 0.026          & 0.027          \\ \hline
Subject6  & 0.024 & 0.025          & \textbf{0.024} & 0.025          & 0.025          & 0.025          & 0.025          & 0.025          & 0.025          \\ \hline
Subject7  & 0.024 & \textbf{0.024} & \textbf{0.024} & \textbf{0.023} & \textbf{0.023} & \textbf{0.024} & \textbf{0.024} & \textbf{0.024} & 0.025          \\ \hline
Subject8  & 0.027 & 0.031          & 0.035          & 0.034          & 0.034          & 0.034          & 0.033          & 0.031          & 0.034          \\ \hline
Subject9  & 0.029 & \textbf{0.029} & 0.030          & 0.030          & 0.030          & 0.030          & 0.037          & 0.037          & 0.038          \\ \hline
Subject10 & 0.033 & \textbf{0.033} & \textbf{0.033} & \textbf{0.033} & \textbf{0.032} & \textbf{0.032} & \textbf{0.032} & \textbf{0.032} & 0.034          \\ \hline
Subject11 & 0.017 & \textbf{0.017} & \textbf{0.016} & \textbf{0.016} & \textbf{0.016} & \textbf{0.016} & \textbf{0.016} & \textbf{0.017} & \textbf{0.017} \\ \hline
Subject12 & 0.019 & \textbf{0.018} & \textbf{0.017} & 0.021          & 0.021          & 0.021          & 0.020          & 0.020          & 0.021          \\ \hline
Subject13 & 0.031 & \textbf{0.030} & \textbf{0.031} & \textbf{0.030} & 0.032          & 0.033          & 0.033          & 0.033          & 0.035          \\ \hline
Subject14 & 0.024 & \textbf{0.020} & \textbf{0.019} & \textbf{0.019} & \textbf{0.019} & \textbf{0.020} & \textbf{0.020} & \textbf{0.020} & 0.027          \\ \hline
Subject15 & 0.019 & 0.020          & \textbf{0.018} & \textbf{0.018} & \textbf{0.018} & \textbf{0.018} & 0.021          & 0.021          & 0.022          \\ \hline
\hline
\end{tabular}
\label{Parkinson_LR}
\end{table}

\begin{table}[h]
\caption{M5P algorithm performance for Parkinson's disease data }
\scriptsize
\begin{tabular}{c|c|cccccccc}
\hline
                            & \textbf{Original data} & \multicolumn{8}{c}{\textbf{After feature selection}}                                                                                                                                                                                                            \\ \hline
\textbf{Number of features} & \textbf{18}            & \multicolumn{1}{c|}{\textbf{11}} & \multicolumn{1}{c|}{\textbf{10}} & \multicolumn{1}{c|}{\textbf{9}} & \multicolumn{1}{c|}{\textbf{8}} & \multicolumn{1}{c|}{\textbf{7}} & \multicolumn{1}{c|}{\textbf{6}} & \multicolumn{1}{c|}{\textbf{5}} & \textbf{4}      \\ \hline
Subject1  & 0.030 & \textbf{0.028} & \textbf{0.028} & \textbf{0.028} & \textbf{0.028} & \textbf{0.028} & \textbf{0.029} & \textbf{0.029} & \textbf{0.029} \\ \hline
Subject2  & 0.027 & 0.028          & 0.029          & 0.030          & 0.030          & 0.030          & 0.030          & 0.030          & 0.030          \\ \hline
Subject3  & 0.018 & 0.019          & \textbf{0.018} & 0.020          & 0.020          & 0.020          & 0.022          & 0.021          & 0.021          \\ \hline
Subject4  & 0.028 & \textbf{0.027} & \textbf{0.027} & \textbf{0.027} & \textbf{0.025} & \textbf{0.027} & \textbf{0.027} & \textbf{0.027} & \textbf{0.028} \\ \hline
Subject5  & 0.024 & \textbf{0.023} & \textbf{0.023} & \textbf{0.023} & \textbf{0.023} & \textbf{0.022} & \textbf{0.022} & \textbf{0.023} & \textbf{0.024} \\ \hline
Subject6  & 0.025 & \textbf{0.025} & \textbf{0.024} & \textbf{0.025} & \textbf{0.025} & \textbf{0.025} & \textbf{0.025} & \textbf{0.025} & \textbf{0.025} \\ \hline
Subject7  & 0.024 & \textbf{0.024} & \textbf{0.024} & \textbf{0.023} & \textbf{0.023} & \textbf{0.024} & \textbf{0.023} & \textbf{0.024} & 0.025          \\ \hline
Subject8  & 0.024 & 0.026          & 0.029          & 0.029          & 0.030          & 0.030          & 0.029          & 0.030          & 0.030          \\ \hline
Subject9  & 0.029 & \textbf{0.029} & \textbf{0.028} & \textbf{0.028} & \textbf{0.029} & \textbf{0.029} & 0.031          & 0.031          & 0.031          \\ \hline
Subject10 & 0.034 & \textbf{0.034} & \textbf{0.034} & \textbf{0.034} & \textbf{0.032} & \textbf{0.032} & \textbf{0.032} & \textbf{0.032} & \textbf{0.032} \\ \hline
Subject11 & 0.017 & \textbf{0.017} & \textbf{0.016} & \textbf{0.016} & \textbf{0.016} & \textbf{0.016} & \textbf{0.016} & \textbf{0.017} & \textbf{0.017} \\ \hline
Subject12 & 0.020 & \textbf{0.018} & \textbf{0.017} & 0.021          & 0.021          & 0.021          & \textbf{0.020} & \textbf{0.020} & 0.021          \\ \hline
Subject13 & 0.033 & \textbf{0.031} & \textbf{0.031} & \textbf{0.031} & \textbf{0.032} & \textbf{0.033} & \textbf{0.033} & \textbf{0.033} & 0.035          \\ \hline
Subject14 & 0.019 & \textbf{0.019} & \textbf{0.019} & \textbf{0.019} & \textbf{0.019} & 0.021          & 0.020          & 0.020          & 0.023          \\ \hline
Subject15 & 0.019 & 0.021          & \textbf{0.019} & \textbf{0.019} & \textbf{0.019} & \textbf{0.019} & 0.023          & 0.023          & 0.022          \\ \hline
\end{tabular}
\label{Parkinson_M5P}
\end{table}


\begin{table}[h]
\caption{M5Rules algorithm performance for Parkinson's disease data}
\scriptsize
\begin{tabular}{c|c|cccccccc}
\hline
                            & \textbf{Original data} & \multicolumn{8}{c}{\textbf{After feature selection}}                                                                                          \\ \hline
\textbf{Number of features} & \textbf{18}            & \textbf{11}     & \textbf{10}     & \textbf{9}      & \textbf{8}      & \textbf{7}      & \textbf{6}      & \textbf{5}      & \textbf{4}      \\ \hline
Subject1  & 0.030 & \textbf{0.028} & \textbf{0.029} & \textbf{0.029} & \textbf{0.028} & \textbf{0.028} & \textbf{0.029} & \textbf{0.029} & \textbf{0.029} \\ \hline
Subject2  & 0.027 & 0.029          & 0.029          & 0.031          & 0.030          & 0.030          & 0.030          & 0.030          & 0.030          \\ \hline
Subject3  & 0.019 & \textbf{0.019} & \textbf{0.018} & 0.020          & 0.020          & 0.021          & 0.022          & 0.021          & 0.021          \\ \hline
Subject4  & 0.029 & \textbf{0.028} & \textbf{0.028} & \textbf{0.027} & \textbf{0.026} & \textbf{0.028} & 0.030          & 0.030          & \textbf{0.028} \\ \hline
Subject5  & 0.025 & \textbf{0.023} & \textbf{0.023} & \textbf{0.023} & \textbf{0.023} & \textbf{0.022} & \textbf{0.022} & \textbf{0.023} & \textbf{0.024} \\ \hline
Subject6  & 0.024 & 0.025          & \textbf{0.024} & 0.025          & 0.025          & 0.025          & 0.025          & 0.025          & 0.025          \\ \hline
Subject7  & 0.024 & \textbf{0.024} & \textbf{0.024} & \textbf{0.023} & \textbf{0.023} & 0.025          & \textbf{0.024} & \textbf{0.024} & 0.025          \\ \hline
Subject8  & 0.029 & 0.038          & 0.043          & 0.043          & 0.031          & 0.031          & 0.044          & 0.045          & 0.031          \\ \hline
Subject9  & 0.031 & \textbf{0.030} & \textbf{0.029} & \textbf{0.029} & \textbf{0.030} & \textbf{0.031} & 0.032          & 0.032          & \textbf{0.031} \\ \hline
Subject10 & 0.035 & \textbf{0.035} & \textbf{0.035} & \textbf{0.034} & \textbf{0.033} & \textbf{0.033} & \textbf{0.033} & \textbf{0.033} & \textbf{0.033} \\ \hline
Subject11 & 0.017 & \textbf{0.017} & \textbf{0.016} & \textbf{0.016} & \textbf{0.016} & \textbf{0.016} & \textbf{0.016} & \textbf{0.017} & \textbf{0.017} \\ \hline
Subject12 & 0.019 & \textbf{0.018} & \textbf{0.017} & 0.021          & 0.021          & 0.021          & 0.020          & 0.020          & 0.021          \\ \hline
Subject13 & 0.033 & \textbf{0.031} & \textbf{0.031} & \textbf{0.031} & \textbf{0.032} & \textbf{0.033} & \textbf{0.033} & \textbf{0.033} & 0.035          \\ \hline
Subject14 & 0.019 & 0.020          & 0.020          & \textbf{0.019} & \textbf{0.019} & 0.021          & 0.020          & 0.020          & 0.023          \\ \hline
Subject15 & 0.020 & 0.021          & \textbf{0.020} & \textbf{0.020} & \textbf{0.020} & \textbf{0.020} & 0.023          & 0.024          & 0.022          \\ \hline
\end{tabular}
\label{Parkinson_M5Rulles}
\end{table}

\begin{figure}
  \centering
       \includegraphics[width=0.45\textwidth]{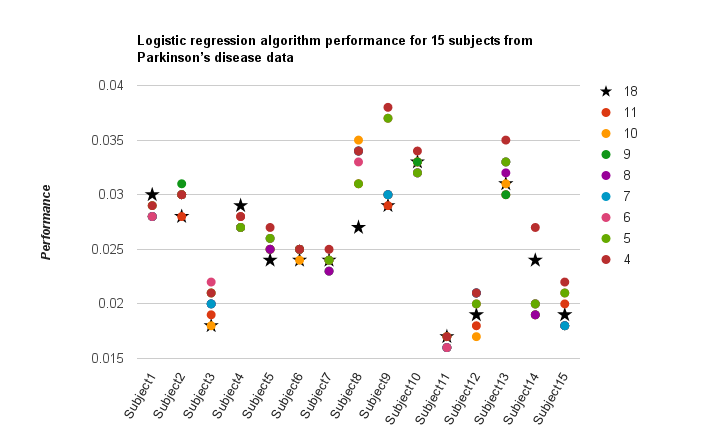}
       \includegraphics[width=0.45\textwidth]{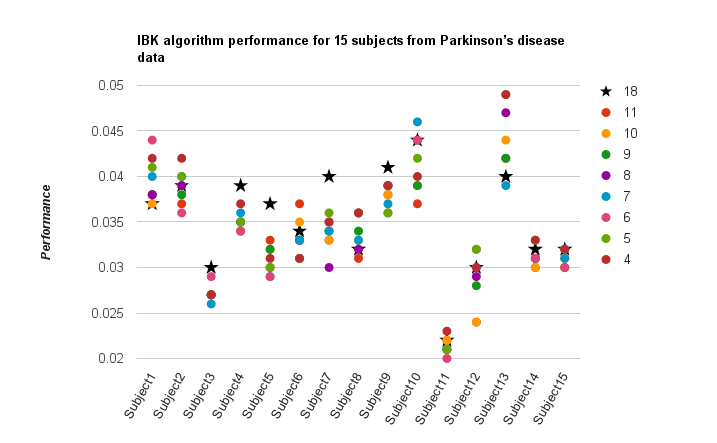}
       \includegraphics[width=0.45\textwidth]{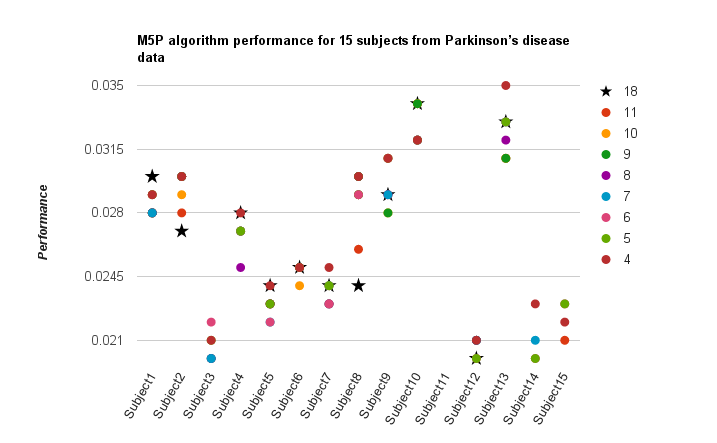}
       \includegraphics[width=0.45\textwidth]{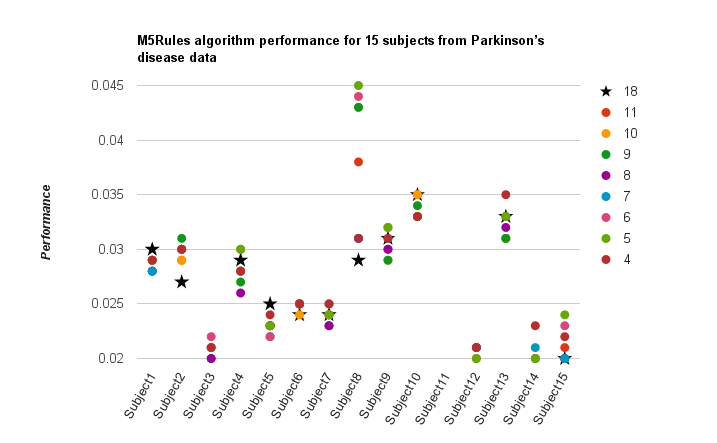}
  \caption{Classification algorithms performance for Parkinson's disease data using all features and after feature selection using cusp catastrophe feature selection algorithm}
  \label{Feature_Selection_Catastrophe_Parkinson}
\end{figure}

Figure \ref{Feature_Selection_Catastrophe_Parkinson} demonstrates applying different classifiers for Parkinson's disease data set. Figure \ref{Feature_Selection_Catastrophe_Parkinson} indicates that cusp model is reduced the error of classifiers for almost all subjects from Parkinson's disease data set.

\begin{figure}[h]
  \centering
       \includegraphics[width=0.9\textwidth]{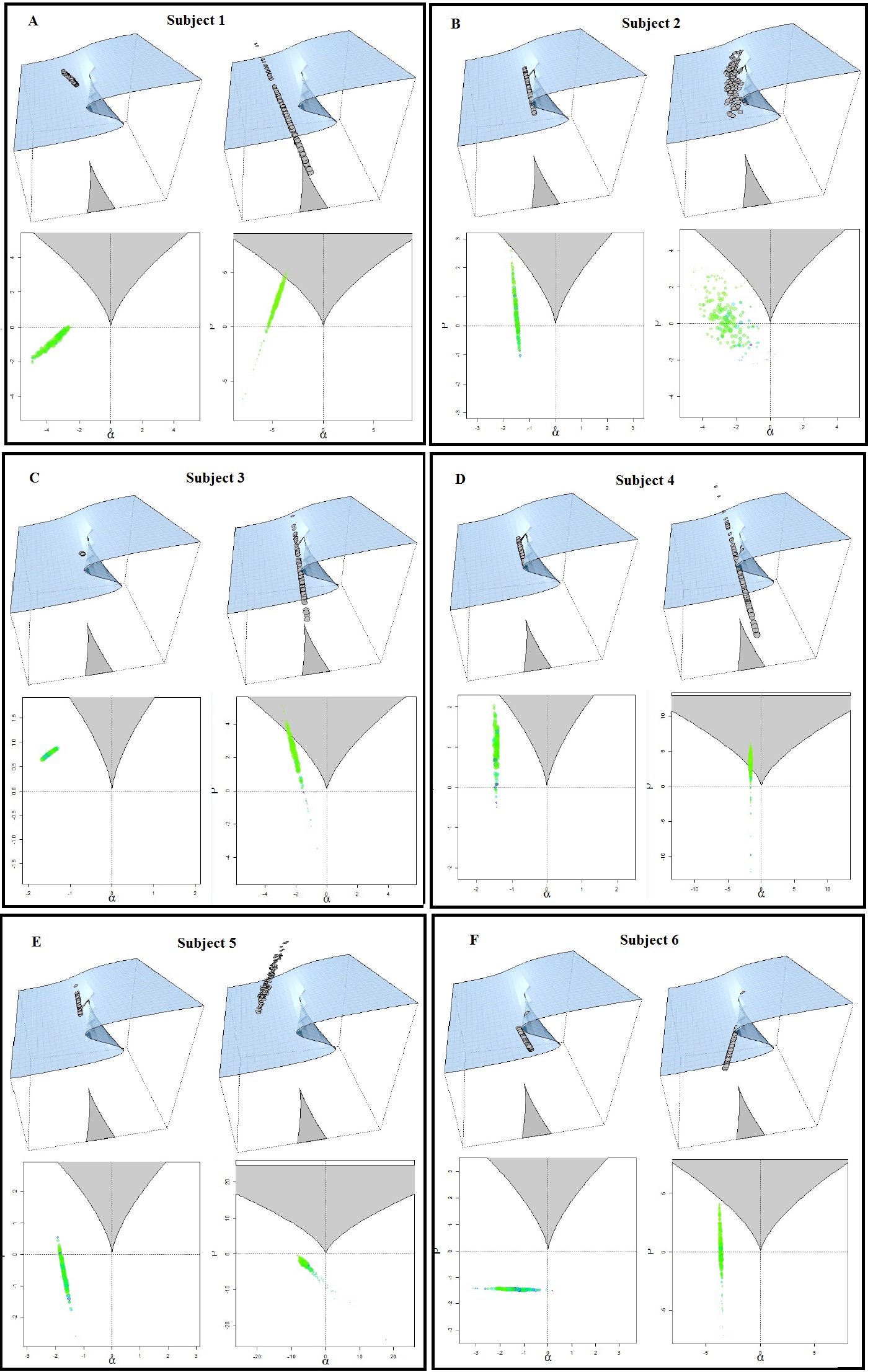}
   \caption{Cusp plot: The most least informative features (left) and the most least informative features (right) base on proposed algorithm for subject 1 to subject 6}
  \label{Cusp_All_Subject}
\end{figure}

Figures \ref{Cusp_All_Subject} show the Equilibrium surface (3 dimensional) and control surface (2 dimensional) of fitting the most irrelevant (left) and the most significant features in different data sets using the Cusp Catastrophe model. The informative features have more affect on the system and put the system closer to the bifurcation situation. 

Tables \ref{rank_S1}- \ref{rank_S6} show the ranking of the features using the proposed and RELIEF algorithms. The ranking values are not exactly the same, but the for almost all cases the informative features' levels are similar in both ranking results. For example, for the first subject, the informative features of 3, 14, 4 and 6 are in the top of the table in both algorithms and less-significant features 2 and 17 are at the bottom.

\begin{table}[h]
\caption{Ranking of the features using the proposed and RELIEF algorithms for subject 1 from Parkinsons disease data}
\centering
\begin{tabular}{|c|c|c|c|}
\hline
\multicolumn{2}{|c|}{Feature selection algorithm based on the Cusp model} & \multicolumn{2}{c|}{RELIEF algorithms} \\ \hline
Attribute ID          & Rank              & Attribute ID          & Rank              \\ \hline
3                     & 0.003144          & 14                    & 0.030901          \\ \hline
14                    & 0.003096          & 3                     & 0.014302          \\ \hline
4                     & 0.003052          & 6                     & 0.014158          \\ \hline
6                     & 0.002947          & 4                     & 0.011554          \\ \hline
15                    & 0.002923          & 5                     & 0.009576          \\ \hline
5                     & 0.002732          & 7                     & 0.009572          \\ \hline
7                     & 0.002731          & 15                    & 0.006487          \\ \hline
9                     & 0.002685          & 12                    & 0.004949          \\ \hline
12                    & 0.002586          & 16                    & 0.004764          \\ \hline
16                    & 0.002569          & 9                     & 0.004034          \\ \hline
8                     & 0.002565          & 11                    & 0.001722          \\ \hline
11                    & 0.002564          & 13                    & 0.0016            \\ \hline
10                    & 0.0025            & 10                    & 0.001595          \\ \hline
13                    & 0.0025            & 2                     & 0.000525          \\ \hline
17                    & 0.002358          & 8                     & -0.00004          \\ \hline
2                     & 0.002351          & 1                     & -0.00254          \\ \hline
1                     & 0.002351          & 17                    & -0.00378          \\ \hline
\end{tabular}
\label{rank_S1}
\end{table}

\begin{table}[h]
\caption{Ranking of the features using the proposed and RELIEF algorithms for subject 2 from Parkinsons disease data}
\centering
\begin{tabular}{|c|c|c|c|}
\hline
\multicolumn{2}{|c|}{Feature selection algorithm based on the Cusp model} & \multicolumn{2}{c|}{RELIEF algorithms} \\ \hline
Attribute ID          & Rank              & Attribute ID          & Rank              \\ \hline
16                    & 0.003336          & 14                    & 0.0195            \\ \hline
6                     & 0.003299          & 6                     & 0.00943           \\ \hline
14                    & 0.003241          & 16                    & 0.00907           \\ \hline
9                     & 0.003212          & 12                    & 0.00784           \\ \hline
12                    & 0.0032            & 9                     & 0.00706           \\ \hline
4                     & 0.003173          & 3                     & 0.00582           \\ \hline
3                     & 0.003167          & 4                     & 0.00309           \\ \hline
15                    & 0.003154          & 15                    & 0.003             \\ \hline
8                     & 0.003134          & 11                    & 0.0026            \\ \hline
13                    & 0.003069          & 13                    & 0.00244           \\ \hline
10                    & 0.003069          & 10                    & 0.00244           \\ \hline
11                    & 0.003057          & 7                     & 0.00243           \\ \hline
7                     & 0.00304           & 5                     & 0.00242           \\ \hline
5                     & 0.00304           & 8                     & 0.0023            \\ \hline
2                     & 0.002727          & 2                     & 0.00133           \\ \hline
1                     & 0.002727          & 1                     & 0.00107           \\ \hline
17                    & 0.002718          & 17                    & -0.00237          \\ \hline
\end{tabular}
\label{rank_S2}
\end{table}

\begin{table}[h]
\caption{Ranking of the features using the proposed and RELIEF algorithms for subject 3 from Parkinsons disease data}
\centering
\begin{tabular}{|c|c|c|c|}
\hline
\multicolumn{2}{|c|}{Feature selection algorithm based on the Cusp model} & \multicolumn{2}{c|}{RELIEF algorithms} \\ \hline
Attribute ID          & Rank              & Attribute ID          & Rank              \\ \hline
15                    & 0.003585          & 15                    & 0.024669          \\ \hline
6                     & 0.003473          & 14                    & 0.018446          \\ \hline
3                     & 0.003261          & 6                     & 0.016579          \\ \hline
4                     & 0.003031          & 3                     & 0.013203          \\ \hline
14                    & 0.002994          & 4                     & 0.010286          \\ \hline
7                     & 0.002946          & 5                     & 0.007498          \\ \hline
9                     & 0.002946          & 7                     & 0.00748           \\ \hline
5                     & 0.002945          & 11                    & 0.005778          \\ \hline
12                    & 0.002937          & 12                    & 0.003904          \\ \hline
8                     & 0.002934          & 9                     & 0.00329           \\ \hline
11                    & 0.002933          & 1                     & 0.003219          \\ \hline
10                    & 0.00287           & 8                     & 0.002655          \\ \hline
13                    & 0.002869          & 10                    & 0.002304          \\ \hline
16                    & 0.002627          & 13                    & 0.002297          \\ \hline
1                     & 0.002595          & 17                    & 0.002161          \\ \hline
2                     & 0.002589          & 2                     & 0.000729          \\ \hline
17                    & 0.002565          & 16                    & -0.00093          \\ \hline
\end{tabular}
\label{rank_S3}
\end{table}

\begin{table}[h]
\caption{Ranking of the features using the proposed and RELIEF algorithms for subject 4 from Parkinsons disease data}
\centering
\begin{tabular}{|c|c|c|c|}
\hline
\multicolumn{2}{|c|}{Feature selection algorithm based on the Cusp model} & \multicolumn{2}{c|}{RELIEF algorithms} \\ \hline
Attribute ID          & Rank              & Attribute ID           & Rank             \\ \hline
3                     & 0.004621          & 6                      & 0.02566          \\ \hline
4                     & 0.00456           & 3                      & 0.02124          \\ \hline
6                     & 0.004473          & 17                     & 0.01921          \\ \hline
5                     & 0.003827          & 4                      & 0.01823          \\ \hline
7                     & 0.003826          & 14                     & 0.01734          \\ \hline
14                    & 0.003417          & 5                      & 0.01714          \\ \hline
15                    & 0.003254          & 7                      & 0.01711          \\ \hline
9                     & 0.002984          & 2                      & 0.00843          \\ \hline
8                     & 0.002968          & 15                     & 0.00774          \\ \hline
13                    & 0.002935          & 13                     & 0.00711          \\ \hline
10                    & 0.002935          & 10                     & 0.00711          \\ \hline
12                    & 0.00293           & 11                     & 0.00695          \\ \hline
11                    & 0.00291           & 12                     & 0.00676          \\ \hline
17                    & 0.002904          & 8                      & 0.00671          \\ \hline
16                    & 0.002793          & 9                      & 0.00613          \\ \hline
1                     & 0.002771          & 1                      & 0.00519          \\ \hline
2                     & 0.00277           & 16                     & 0.00168          \\ \hline
\end{tabular}
\label{rank_S4}
\end{table}

\begin{table}[h]
\caption{Ranking of the features using the proposed and RELIEF algorithms for subject 5 from Parkinsons disease data}
\centering
\begin{tabular}{|c|c|c|c|}
\hline
\multicolumn{2}{|c|}{Feature selection algorithm based on the Cusp model} & \multicolumn{2}{c|}{RELIEF algorithms} \\ \hline
Attribute ID          & Rank              & Attribute ID          & Rank              \\ \hline
14                    & 0.003896          & 14                    & 0.02979           \\ \hline
3                     & 0.003671          & 6                     & 0.02661           \\ \hline
4                     & 0.003533          & 4                     & 0.02327           \\ \hline
6                     & 0.003529          & 3                     & 0.01819           \\ \hline
7                     & 0.003189          & 7                     & 0.01289           \\ \hline
5                     & 0.003185          & 5                     & 0.01287           \\ \hline
15                    & 0.003059          & 9                     & 0.01101           \\ \hline
16                    & 0.00253           & 15                    & 0.0101            \\ \hline
9                     & 0.00248           & 12                    & 0.00659           \\ \hline
12                    & 0.002401          & 11                    & 0.00414           \\ \hline
8                     & 0.002372          & 10                    & 0.00354           \\ \hline
11                    & 0.002363          & 13                    & 0.00354           \\ \hline
10                    & 0.002343          & 8                     & 0.00331           \\ \hline
13                    & 0.002343          & 16                    & 0.00278           \\ \hline
2                     & 0.002339          & 2                     & 0.00244           \\ \hline
1                     & 0.002324          & 17                    & 0.00116           \\ \hline
17                    & 0.002314          & 1                     & -0.00406          \\ \hline
\end{tabular}
\label{rank_S5}
\end{table}

\begin{table}[h]
\caption{Ranking of the features using the proposed and RELIEF algorithms for subject 6 from Parkinsons disease data}
\centering
\begin{tabular}{|c|c|c|c|}
\hline
\multicolumn{2}{|c|}{Feature selection algorithm based on the Cusp model} & \multicolumn{2}{c|}{RELIEF algorithms} \\ \hline
Attribute ID          & Rank              & Attribute ID          & Rank              \\ \hline
15                    & 0.003297          & 14                    & 0.014851          \\ \hline
4                     & 0.003173          & 6                     & 0.014336          \\ \hline
3                     & 0.003093          & 15                    & 0.014142          \\ \hline
6                     & 0.003076          & 17                    & 0.01388           \\ \hline
14                    & 0.003062          & 4                     & 0.012454          \\ \hline
7                     & 0.002854          & 3                     & 0.010648          \\ \hline
5                     & 0.002854          & 7                     & 0.008541          \\ \hline
9                     & 0.002691          & 5                     & 0.008525          \\ \hline
12                    & 0.002649          & 2                     & 0.003976          \\ \hline
8                     & 0.002644          & 12                    & 0.002502          \\ \hline
11                    & 0.002619          & 1                     & 0.001973          \\ \hline
16                    & 0.002597          & 11                    & 0.001794          \\ \hline
10                    & 0.002565          & 9                     & -6.8E-05          \\ \hline
13                    & 0.002565          & 8                     & -0.0006           \\ \hline
1                     & 0.002418          & 13                    & -0.00153          \\ \hline
2                     & 0.00234           & 10                    & -0.00153          \\ \hline
17                    & 0.002315          & 16                    & -0.00244          \\ \hline
\end{tabular}
\label{rank_S6}
\end{table}

Tables \ref{MAE_LR}-\ref{RMSE_REPTree} show the mean absolute error and root mean square error for Regression analysis before and after feature selection for 15 subjects. We separated the results of different algorithms from each other. Tables \ref{MAE_LR} and \ref{RMSE_LR} shows the results of Linear regression algorithm. The accuracy of analyzing all subjects except subject 2, 9 and 14 using the proposed algorithm compared with original data is improved. The RELIEF algorithm has improvement for almost all subjects, but our algorithm has better performance than RELIEF algorithm. \par
Tables \ref{MAE_IBK}-\ref{RMSE_IBK} are the related results for K-nearest neighbors algorithm and they show that both algorithms have better accuracy only for 60\% of subjects and the same situation happened for M5Rulles (see the tables \ref{MAE_M5Rulles}-\ref{RMSE_M5Rulles}) and REPTree (\ref{MAE_REPTree}-\ref{RMSE_REPTree}) algorithms, but for some subjects the RELIEF algorithm has better performance.

\begin{table}[h]
\caption{Mean absolute error of Linear regression algorithm after feature selection using the proposed and RELIEF algorithms for Slice locality data set}
\centering
\small
\begin{tabular}{|c|c|c|c|}
\hline
        & \multicolumn{3}{c|}{MAE of Linear Regression}                                                                                                                                                                \\ \hline
Subject & Original data & \begin{tabular}[c]{@{}c@{}}After Feature selection\\   using feature selection algorithm\\ based on the Cusp model\end{tabular} & \begin{tabular}[c]{@{}c@{}}After feature\\   selection using RELIEF algorithms\end{tabular} \\ \hline
1       & 0.0295        & 0.0291                                                                                      & 0.0282                                                                                        \\ \hline
2       & 0.0276        & 0.028                                                                                       & 0.028                                                                                         \\ \hline
3       & 0.0183        & 0.0183                                                                                      & 0.0182                                                                                        \\ \hline
4       & 0.0292        & 0.029                                                                                       & 0.0292                                                                                        \\ \hline
5       & 0.0235        & 0.0235                                                                                      & 0.0235                                                                                        \\ \hline
6       & 0.0239        & 0.0239                                                                                      & 0.024                                                                                         \\ \hline
7       & 0.0243        & 0.0242                                                                                      & 0.0244                                                                                        \\ \hline
8       & 0.0266        & 0.0266                                                                                      & 0.028                                                                                         \\ \hline
9       & 0.0286        & 0.0288                                                                                      & 0.0288                                                                                        \\ \hline
10      & 0.0333        & 0.0333                                                                                      & 0.0333                                                                                        \\ \hline
11      & 0.0169        & 0.0167                                                                                      & 0.017                                                                                         \\ \hline
12      & 0.0193        & 0.0187                                                                                      & 0.0194                                                                                        \\ \hline
13      & 0.0305        & 0.0297                                                                                      & 0.0315                                                                                        \\ \hline
14      & 0.019         & 0.0193                                                                                      & 0.0188                                                                                        \\ \hline
15      & 0.0266        & 0.0261                                                                                      & 0.0266                                                                                        \\ \hline
\end{tabular}
\label{MAE_LR}
\end{table}

\begin{table}[h]
\caption{Root mean square error of Linear regression algorithm after feature selection using the proposed and RELIEF algorithms for Slice locality data set}
\small
\centering
\begin{tabular}{|c|c|c|c|}
\hline
        & \multicolumn{3}{c|}{RMSE of Linear Regression}                                                                                                                                                                \\ \hline
Subject & Original data & \begin{tabular}[c]{@{}c@{}}After Feature selection\\   using feature selection algorithm\\ based on the Cusp model\end{tabular} & \begin{tabular}[c]{@{}c@{}}After feature\\   selection using RELIEF algorithm\end{tabular} \\ \hline
1       & 0.0386        & 0.0381                                                                                      & 0.0384                                                                                        \\ \hline
2       & 0.0372        & 0.0377                                                                                      & 0.0377                                                                                        \\ \hline
3       & 0.0249        & 0.0249                                                                                      & 0.0248                                                                                        \\ \hline
4       & 0.042         & 0.0418                                                                                      & 0.042                                                                                         \\ \hline
5       & 0.0336        & 0.0336                                                                                      & 0.0336                                                                                        \\ \hline
6       & 0.0325        & 0.0325                                                                                      & 0.0322                                                                                        \\ \hline
7       & 0.0338        & 0.0338                                                                                      & 0.0335                                                                                        \\ \hline
8       & 0.0401        & 0.0401                                                                                      & 0.0424                                                                                        \\ \hline
9       & 0.0376        & 0.0377                                                                                      & 0.0375                                                                                        \\ \hline
10      & 0.0472        & 0.0472                                                                                      & 0.0461                                                                                        \\ \hline
11      & 0.0239        & 0.0237                                                                                      & 0.024                                                                                         \\ \hline
12      & 0.025         & 0.0245                                                                                      & 0.0256                                                                                        \\ \hline
13      & 0.0404        & 0.0392                                                                                      & 0.0425                                                                                        \\ \hline
14      & 0.0248        & 0.0253                                                                                      & 0.0246                                                                                        \\ \hline
15      & 0.0322        & 0.0317                                                                                      & 0.0319                                                                                        \\ \hline
\end{tabular}
\label{RMSE_LR}
\end{table}

\begin{table}[h]
\caption{Mean absolute error of IBK algorithm after feature selection using the proposed and RELIEF algorithms for Slice locality data set}
\small
\centering
\begin{tabular}{|c|c|c|c|}
\hline
        & \multicolumn{3}{c|}{MAE of IBK}                                                                                                                                                               \\ \hline
Subject & Original data & \begin{tabular}[c]{@{}c@{}}After Feature selection\\   using feature selection algorithm\\ based on the Cusp model\end{tabular} & \begin{tabular}[c]{@{}c@{}}After feature\\   selection using RELIEF algorithm\end{tabular} \\ \hline
1       & 0.037         & 0.038                                                                                       & 0.042                                                                                         \\ \hline
2       & 0.0389        & 0.0411                                                                                      & 0.0411                                                                                        \\ \hline
3       & 0.0304        & 0.0297                                                                                      & 0.0311                                                                                        \\ \hline
4       & 0.0394        & 0.0387                                                                                      & 0.0372                                                                                        \\ \hline
5       & 0.0369        & 0.0344                                                                                      & 0.0356                                                                                        \\ \hline
6       & 0.034         & 0.0334                                                                                      & 0.0355                                                                                        \\ \hline
7       & 0.0404        & 0.0385                                                                                      & 0.0389                                                                                        \\ \hline
8       & 0.0321        & 0.032                                                                                       & 0.032                                                                                         \\ \hline
9       & 0.0405        & 0.0399                                                                                      & 0.0399                                                                                        \\ \hline
10      & 0.0439        & 0.044                                                                                       & 0.0433                                                                                        \\ \hline
11      & 0.0218        & 0.0231                                                                                      & 0.0224                                                                                        \\ \hline
12      & 0.0297        & 0.0295                                                                                      & 0.0308                                                                                        \\ \hline
13      & 0.0402        & 0.0411                                                                                      & 0.0402                                                                                        \\ \hline
14      & 0.0317        & 0.03                                                                                        & 0.0307                                                                                        \\ \hline
15      & 0.0338        & 0.0352                                                                                      & 0.0335                                                                                        \\ \hline
\end{tabular}
\label{MAE_IBK}
\end{table}

\begin{table}[h]
\caption{Root mean square error of IBK algorithm after feature selection using the proposed and RELIEF algorithms for Slice locality data set}
\small
\centering
\begin{tabular}{|c|c|c|c|}
\hline
        & \multicolumn{3}{c|}{RMSE of IBK}                                                                                                                                                                            \\ \hline
Subject & Original data & \begin{tabular}[c]{@{}c@{}}After Feature selection\\   using feature selection algorithm\\ based on the Cusp model\end{tabular} & \begin{tabular}[c]{@{}c@{}}After feature\\   selection using RELIEF algorithm\end{tabular} \\ \hline
1       & 0.0493        & 0.0506                                                                                      & 0.0548                                                                                        \\ \hline
2       & 0.0526        & 0.0537                                                                                      & 0.0537                                                                                        \\ \hline
3       & 0.0379        & 0.0379                                                                                      & 0.0401                                                                                        \\ \hline
4       & 0.0569        & 0.0565                                                                                      & 0.0567                                                                                        \\ \hline
5       & 0.0499        & 0.047                                                                                       & 0.0477                                                                                        \\ \hline
6       & 0.0453        & 0.0447                                                                                      & 0.0457                                                                                        \\ \hline
7       & 0.0527        & 0.0504                                                                                      & 0.0507                                                                                        \\ \hline
8       & 0.0462        & 0.0458                                                                                      & 0.0466                                                                                        \\ \hline
9       & 0.0531        & 0.0528                                                                                      & 0.0536                                                                                        \\ \hline
10      & 0.056         & 0.0562                                                                                      & 0.0538                                                                                        \\ \hline
11      & 0.0285        & 0.0316                                                                                      & 0.029                                                                                         \\ \hline
12      & 0.0385        & 0.0399                                                                                      & 0.0402                                                                                        \\ \hline
13      & 0.0533        & 0.0539                                                                                      & 0.0532                                                                                        \\ \hline
14      & 0.038         & 0.0359                                                                                      & 0.0378                                                                                        \\ \hline
15      & 0.0426        & 0.0435                                                                                      & 0.0427                                                                                        \\ \hline
\end{tabular}
\label{RMSE_IBK}
\end{table}


\begin{table}[h]
\caption{Mean absolute error of M5Rules algorithm after feature selection using the proposed and RELIEF algorithms for Slice locality data set}
\small
\centering
\begin{tabular}{|c|c|c|c|}
\hline
        & \multicolumn{3}{c|}{MAE of M5Rules}                                                                                                                                                                          \\ \hline
Subject & Original data & \begin{tabular}[c]{@{}c@{}}After Feature selection\\   using feature selection algorithm\\ based on the Cusp model\end{tabular} & \begin{tabular}[c]{@{}c@{}}After feature\\   selection using RELIEF algorithm\end{tabular} \\ \hline
1       & 0.0299        & 0.0299                                                                                      & 0.0292                                                                                        \\ \hline
2       & 0.0273        & 0.0285                                                                                      & 0.0276                                                                                        \\ \hline
3       & 0.0188        & 0.0181                                                                                      & 0.0203                                                                                        \\ \hline
4       & 0.0291        & 0.0291                                                                                      & 0.0278                                                                                        \\ \hline
5       & 0.0246        & 0.0248                                                                                      & 0.0233                                                                                        \\ \hline
6       & 0.0241        & 0.024                                                                                       & 0.024                                                                                         \\ \hline
7       & 0.0237        & 0.0233                                                                                      & 0.0235                                                                                        \\ \hline
8       & 0.0286        & 0.0275                                                                                      & 0.0262                                                                                        \\ \hline
9       & 0.0306        & 0.0319                                                                                      & 0.0306                                                                                        \\ \hline
10      & 0.0349        & 0.0349                                                                                      & 0.034                                                                                         \\ \hline
11      & 0.0167        & 0.0169                                                                                      & 0.0175                                                                                        \\ \hline
12      & 0.019         & 0.0188                                                                                      & 0.0197                                                                                        \\ \hline
13      & 0.0333        & 0.0313                                                                                      & 0.032                                                                                         \\ \hline
14      & 0.0196        & 0.021                                                                                       & 0.0209                                                                                        \\ \hline
15      & 0.0246        & 0.0249                                                                                      & 0.0256                                                                                        \\ \hline
\end{tabular}
\label{MAE_M5Rulles}
\end{table}

\begin{table}[h]
\caption{Root mean square error of M5Rules algorithm after feature selection using the proposed and RELIEF algorithms for Slice locality data set}
\small
\centering
\begin{tabular}{|c|c|c|c|}
\hline
        & \multicolumn{3}{c|}{RMSE of M5Rules}                                                                                                                                                                         \\ \hline
Subject & Original data & \begin{tabular}[c]{@{}c@{}}After Feature selection\\   using feature selection algorithm\\ based on the Cusp model\end{tabular} & \begin{tabular}[c]{@{}c@{}}After feature\\   selection using RELIEF algorithm\end{tabular} \\ \hline
1       & 0.0393        & 0.0393                                                                                      & 0.0393                                                                                        \\ \hline
2       & 0.0366        & 0.0377                                                                                      & 0.0366                                                                                        \\ \hline
3       & 0.0259        & 0.0252                                                                                      & 0.0284                                                                                        \\ \hline
4       & 0.0423        & 0.0423                                                                                      & 0.0415                                                                                        \\ \hline
5       & 0.0343        & 0.0345                                                                                      & 0.0318                                                                                        \\ \hline
6       & 0.0327        & 0.0327                                                                                      & 0.0327                                                                                        \\ \hline
7       & 0.033         & 0.0324                                                                                      & 0.0328                                                                                        \\ \hline
8       & 0.0457        & 0.0444                                                                                      & 0.0388                                                                                        \\ \hline
9       & 0.0403        & 0.0424                                                                                      & 0.0401                                                                                        \\ \hline
10      & 0.0488        & 0.0488                                                                                      & 0.0478                                                                                        \\ \hline
11      & 0.0219        & 0.0225                                                                                      & 0.0235                                                                                        \\ \hline
12      & 0.0244        & 0.0244                                                                                      & 0.026                                                                                         \\ \hline
13      & 0.044         & 0.041                                                                                       & 0.0424                                                                                        \\ \hline
14      & 0.0258        & 0.0286                                                                                      & 0.0275                                                                                        \\ \hline
15      & 0.0305        & 0.0305                                                                                      & 0.0314                                                                                        \\ \hline
\end{tabular}
\label{RMSE_M5Rulles}
\end{table}


\begin{table}[h]
\caption{Mean absolute error of REPTree algorithm after feature selection using the proposed and RELIEF algorithms for Slice locality data set}
\small
\centering
\begin{tabular}{|c|c|c|c|}
\hline
        & \multicolumn{3}{c|}{MAE of REPTree}                                                                                                                                                                        \\ \hline
Subject & Original data & \begin{tabular}[c]{@{}c@{}}After Feature selection\\   using feature selection algorithm\\ based on the Cusp model\end{tabular} & \begin{tabular}[c]{@{}c@{}}After feature\\   selection using RELIEF algorithm\end{tabular} \\ \hline
1       & 0.0357        & 0.0357                                                                                      & 0.0353                                                                                        \\ \hline
2       & 0.0344        & 0.0347                                                                                      & 0.0347                                                                                        \\ \hline
3       & 0.0223        & 0.0228                                                                                      & 0.0226                                                                                        \\ \hline
4       & 0.0312        & 0.0308                                                                                      & 0.0304                                                                                        \\ \hline
5       & 0.0272        & 0.0273                                                                                      & 0.0276                                                                                        \\ \hline
6       & 0.0278        & 0.028                                                                                       & 0.0278                                                                                        \\ \hline
7       & 0.0273        & 0.0276                                                                                      & 0.0276                                                                                        \\ \hline
8       & 0.03          & 0.0311                                                                                      & 0.03                                                                                          \\ \hline
9       & 0.0387        & 0.0381                                                                                      & 0.0387                                                                                        \\ \hline
10      & 0.0358        & 0.0358                                                                                      & 0.0349                                                                                        \\ \hline
11      & 0.0183        & 0.018                                                                                       & 0.0184                                                                                        \\ \hline
12      & 0.0261        & 0.0267                                                                                      & 0.0261                                                                                        \\ \hline
13      & 0.043         & 0.043                                                                                       & 0.0428                                                                                        \\ \hline
14      & 0.0263        & 0.0263                                                                                      & 0.0263                                                                                        \\ \hline
15      & 0.0288        & 0.0289                                                                                      & 0.0293                                                                                        \\ \hline
\end{tabular}
\label{MAE_REPTree}
\end{table}

\begin{table}[h]
\caption{Root mean square error of REPTree algorithm after feature selection using the proposed and RELIEF algorithms for Slice locality data set}
\small
\centering
\begin{tabular}{|c|c|c|c|}
\hline
        & \multicolumn{3}{c|}{RMSE of REPTree}                                                                                                                                                                         \\ \hline
Subject & Original data & \begin{tabular}[c]{@{}c@{}}After Feature selection\\   using feature selection algorithm\\ based on the Cusp model\end{tabular} & \begin{tabular}[c]{@{}c@{}}After feature\\   selection using RELIEF algorithm\end{tabular} \\ \hline
1       & 0.0458        & 0.0458                                                                                      & 0.0453                                                                                        \\ \hline
2       & 0.0449        & 0.0448                                                                                      & 0.0448                                                                                        \\ \hline
3       & 0.0284        & 0.0288                                                                                      & 0.0288                                                                                        \\ \hline
4       & 0.0449        & 0.0446                                                                                      & 0.0437                                                                                        \\ \hline
5       & 0.0363        & 0.0363                                                                                      & 0.0367                                                                                        \\ \hline
6       & 0.0371        & 0.038                                                                                       & 0.0371                                                                                        \\ \hline
7       & 0.0379        & 0.0387                                                                                      & 0.0383                                                                                        \\ \hline
8       & 0.0506        & 0.0547                                                                                      & 0.0506                                                                                        \\ \hline
9       & 0.0519        & 0.0513                                                                                      & 0.052                                                                                         \\ \hline
10      & 0.0458        & 0.0458                                                                                      & 0.0454                                                                                        \\ \hline
11      & 0.0251        & 0.025                                                                                       & 0.0252                                                                                        \\ \hline
12      & 0.0336        & 0.0356                                                                                      & 0.0335                                                                                        \\ \hline
13      & 0.0538        & 0.0538                                                                                      & 0.0538                                                                                        \\ \hline
14      & 0.0339        & 0.0338                                                                                      & 0.0339                                                                                        \\ \hline
15      & 0.0362        & 0.0363                                                                                      & 0.0366                                                                                        \\ \hline
\end{tabular}
\label{RMSE_REPTree}
\end{table}

\begin{figure}
\centering
       \includegraphics[width=0.45\textwidth]{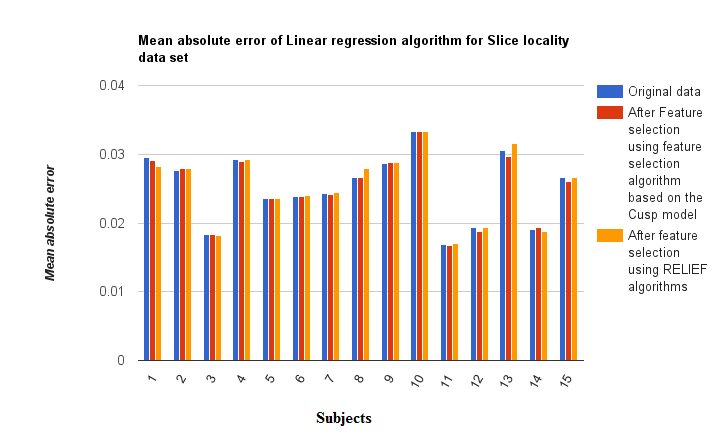}
       \includegraphics[width=0.45\textwidth]{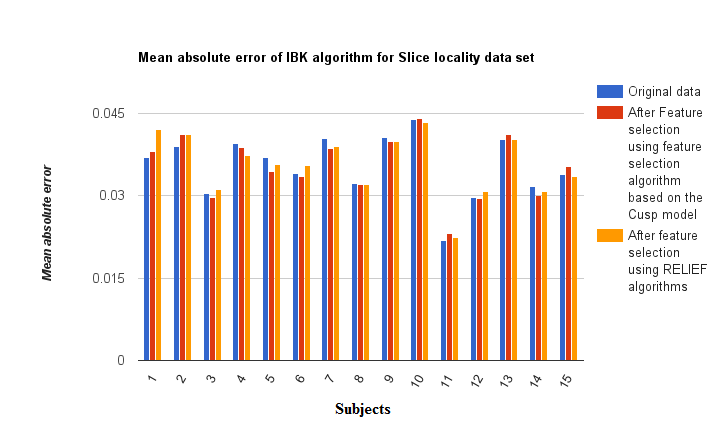}
       \includegraphics[width=0.45\textwidth]{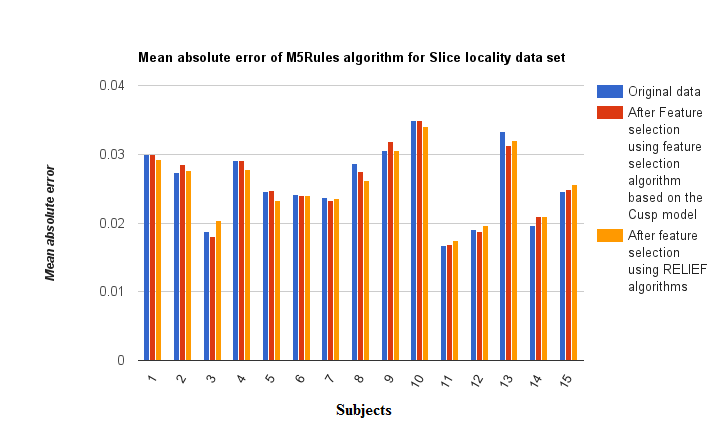}
       \includegraphics[width=0.45\textwidth]{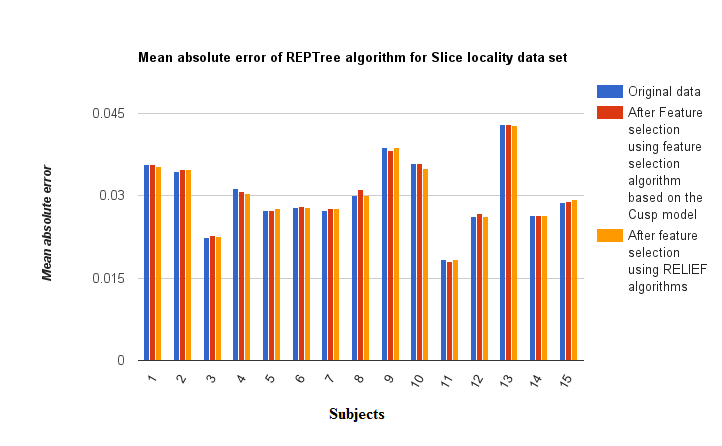}
%
       \includegraphics[width=0.45\textwidth]{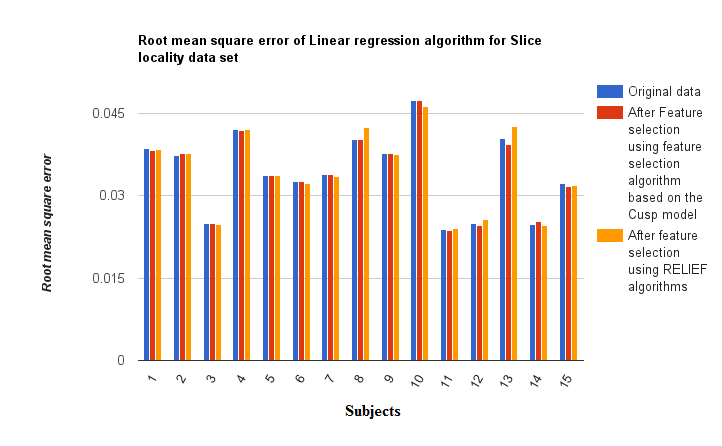}
       \includegraphics[width=0.45\textwidth]{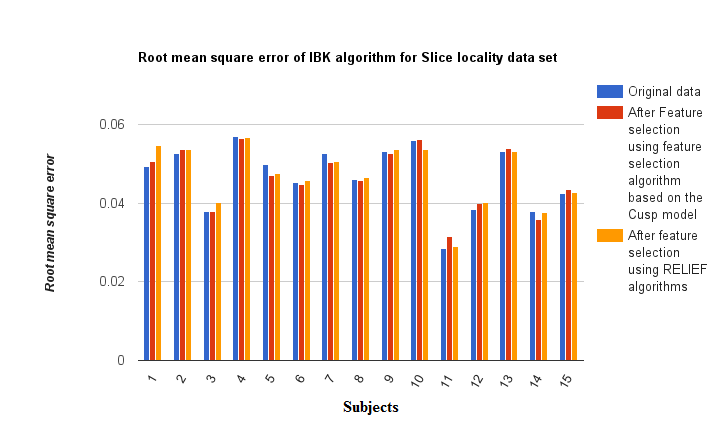}
       \includegraphics[width=0.45\textwidth]{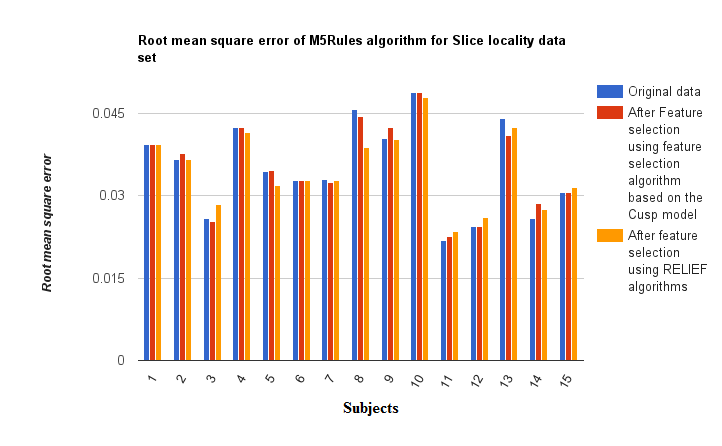}
       \includegraphics[width=0.45\textwidth]{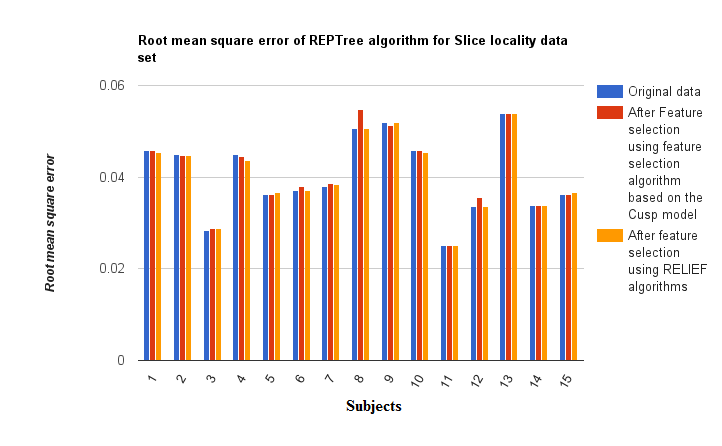}
  \caption{Mean square error and root mean square error of classifiers after feature selection using the proposed and RELIEF algorithms for Slice locality data set}
  \label{Feature_Selection_Catastrophe_MAE_RMSE_Slice}
\end{figure}

Figure \ref{Feature_Selection_Catastrophe_MAE_RMSE_Slice} provides a comparison between proposed algorithm and the well known RELIEF algorithm for Slice locality data set. Mean absolute error and root mean square error of four classifiers of original data and after feature selection are shown in the figures. The graph show that the proposed algorithm is improved the accuracy of classification algorithms for almost all subjects using different classifiers.

\section{Conclusions} \label{conclusions1}
In this paper, we introduced a new feature selection algorithms to remove the irrelevant or redundant features in the data sets. This algorithm removes the irrelevant or redundant features of a regression data sets. This algorithm selects significant features based on their fitting to the Catastrophe model and the features that better change the dynamics of the outcome feature or features are considered as informative features. The Akaike information criterion value of the Cusp model is computed for ranking of each feature. We applied this algorithm to three different data sets: Parkinson's Telemonitoring, Breast Cancer and Slice locality from UCI machine learning repository. Results show that the proposed algorithm is efficient in finding the significant subset of features in a data set.

\clearpage

\bibliographystyle{acm}
\bibliography{database}
\end{document}